\documentclass[journal, onecolumn]{IEEEtran}
\ifCLASSINFOpdf
\else
\fi
\usepackage{graphicx}
\usepackage{comment}
\usepackage{cite}
\usepackage{subfigure}
\usepackage[fleqn]{amsmath}
\usepackage{url}
\usepackage{color}

\begin{document}

\title{Deep Recurrent Neural Networks for mapping winter vegetation quality coverage via multi-temporal SAR Sentinel-1}

\author{Dinh Ho Tong Minh,
		Dino Ienco,
        Raffaele Gaetano,
        Nathalie Lalande,
        Emile Ndikumana,
        Faycal Osman,
        and Pierre Maurel
        
\thanks{ D. Ho Tong Minh, E. Ndikumana, F. Osman and P. Maurel are with UMR-TETIS laboratory, IRSTEA, Montpellier, France (email: dinh.ho-tong-minh; emile.ndikumana; faycal.osman; pierre.maurel@irstea.fr)}
\thanks{D. Ienco is with UMR-TETIS laboratory, IRSTEA, University of Montpellier, Montpellier, France and with LIRMM laboratory, Montpellier, France (email: dino.ienco@irstea.fr).}%
\thanks{R. Gaetano is with UMR-TETIS laboratory, CIRAD, University of Montpellier, Montpellier, France (email: raffaele.gaetano@cirad.fr).}
\thanks{N. Lalande is with Envylis Cie, Villeneuve-les-Maguelone, France (email: nathalie.lalande@envilys.com).}}

\maketitle

\begin{abstract}
Mapping winter vegetation quality coverage is a challenge problem of remote sensing. This is due to the cloud coverage in winter period, leading to use radar  rather than optical images. The objective of this paper is to provide a better understanding of the capabilities of radar Sentinel-1 and deep learning concerning about  mapping winter vegetation quality coverage. The analysis presented in this paper is carried out on multi-temporal Sentinel-1 data over the site of La Rochelle, France, during the campaign in December 2016. This dataset were processed in order to produce an intensity radar data stack from October 2016 to February 2017. Two deep Recurrent Neural Network (RNN) based classifier methods were employed. We found that the results of RNNs clearly outperformed the classical machine learning approaches (Support Vector Machine and Random Forest). This study confirms that the  time series radar Sentinel-1 and RNNs could be exploited for winter vegetation quality cover mapping.
\end{abstract}


\section{Introduction\label{sec:Introduction}}

\IEEEPARstart{ a}{n} excess of nitrates in drinking water is harmful to human health. They come mainly from surplus agricultural fertilizers leached when it rains on bare soil. Since 2011, the European Nitrates Directive has obliged to reduce diffuse pollution in the perimeters of drinking water collection thanks to a winter vegetation covering of the soils  \cite{Kallis01}. It makes it possible to absorb a part of the surpluses, and above all to limit the leaching of the soil. When spring and summer crops release the soil from any vegetation cover in autumn and early winter, farmers have to establish intermediate crops (cereals, grasslands, fallow land, etc.) and receive aid in return \cite{Buckley13}. The mapping of the winter cover in vulnerable areas helps the services of the State to control the declarations of the farmers but also the local actors to set up innovative action plans to reduce diffuse pollution. Remote sensing satellite imagery is a valuable aid in understanding the level of vegetation cover in winter \cite{Hively09}. Moreover, the high cloud cover levels during the winter season and the dynamic of vegetation justify the use of radar images time series rather than optical ones. Recently,  the Sentinel-1 satellite is operating day and night and performing C-band synthetic aperture radar (SAR) imaging, enabling it to acquire imagery regardless of the weather \cite{TORRES20129}. The Sentinel-1 is systematically acquired data at global scale  in Terrain Observation with Progressive Scan (TOPS) mode with a revisit period of 12 days. This offers a unique source of information to map our vegetation quality cover in winter season.   

In the literature of remote sensing for land cover mapping, most of works are based on standard machine learning approach (i.e., Support vector machine (SVM), Random Forest (RF)). These standard approaches however ignore any temporal dependency of a time series data. Recent advances in machine learning, deep learning approaches, have demonstrated their powerful value. In particular, Recurrent Neural Network (RNN) \cite{BengioCV13} is a family of deep learning methods, which offers a certain model (e.g., Long Short Term Memory (LSTM) \cite{HochreiterS96}) to exploit temporal dependency among data and, in our case, the temporal dependency available in a time series SAR Sentinel-1 data.  

In this paper, we aim to provide a better understanding of the capabilities of SAR Sentinel-1 and deep RNN concerning about  mapping winter vegetation quality coverage.

\section{Study area \label{sec:area}}

\subsection{La Rochelle site\label{sec:IRF-1} }

The study area (horizontal: 40 km, vertical: 60 km) is located in the department of Charentes-Maritimes, in the west of France, a very vulnerable zone to diffuse pollution. The Re-Sources actions program (2016-2020) aims to improve water quality while maintaining an efficient agricultural both in environmental and economic terms. It is based on a close partnership between the State services, the Water Agency, the Region and local actors (municipalities, professional agricultural organizations and farmers). 

\subsection{Ground data\label{sec:IRF-2} }
\begin{figure}[ht!]
\centering
\includegraphics[scale=0.25]{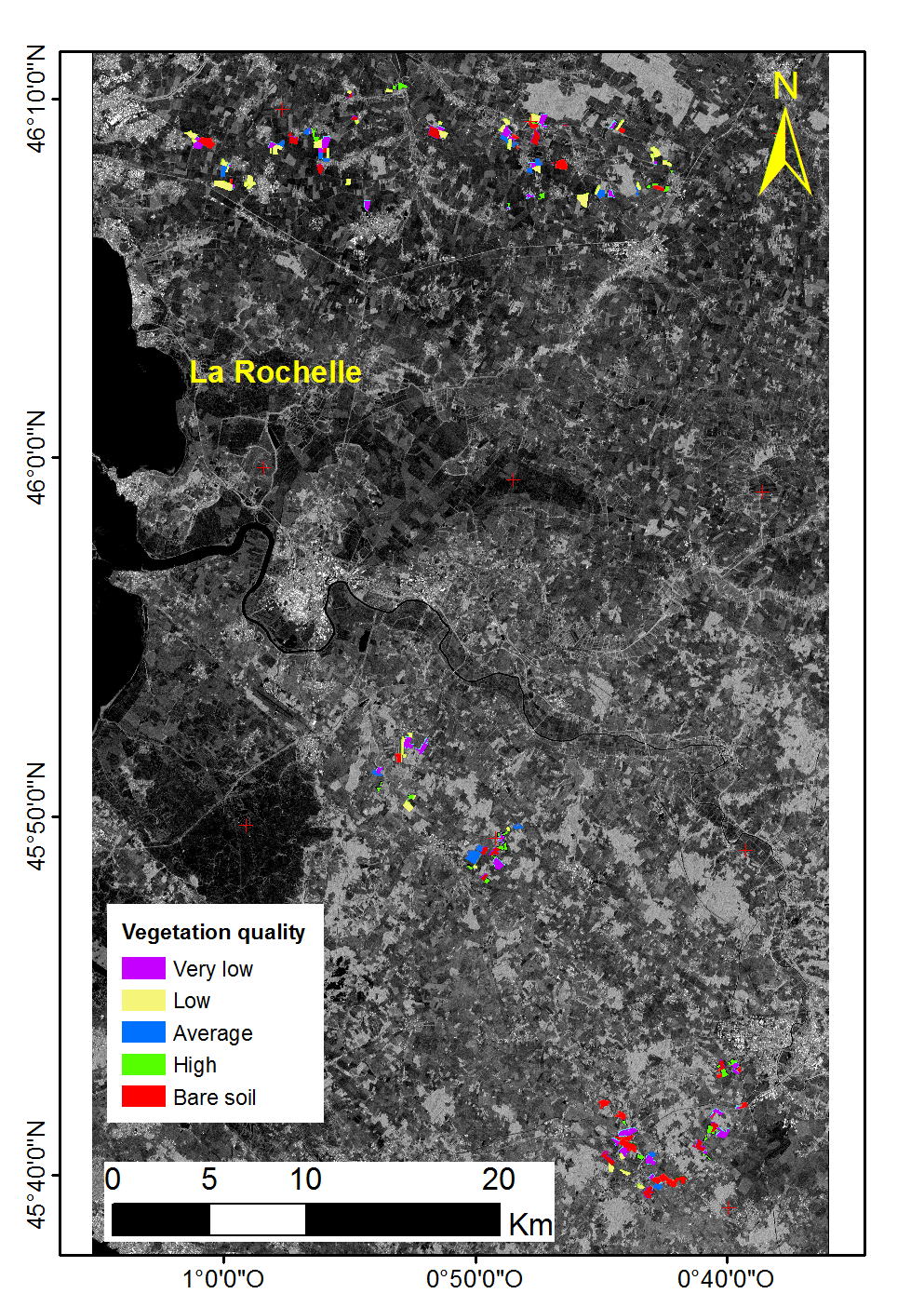}
\caption{The La Rochelle study site  (horizontal: 40 km, vertical: 60 km) is located in the department of Charentes-Maritimes, in the west of France. The background is the average VV intensity. Color polygons represent field locations. \label{fig:StudyArea}}
\end{figure}

A field survey was carried out on 194 plots in December 2016 by the local actors with the help of an environmental engineering company. The following information was collected on each plot: land cover, an estimate of the quality of the vegetative development (5 classes: bare soil, very low, low, average, high), the date of observation and a photograph. Figure \ref{fig:StudyArea} shows the study area and ground  measured data.

\section{SAR data and processing \label{sec:SAR-data}}

\subsection{SAR data }

\begin{figure}[!ht]
\centering
\includegraphics[scale=0.50]{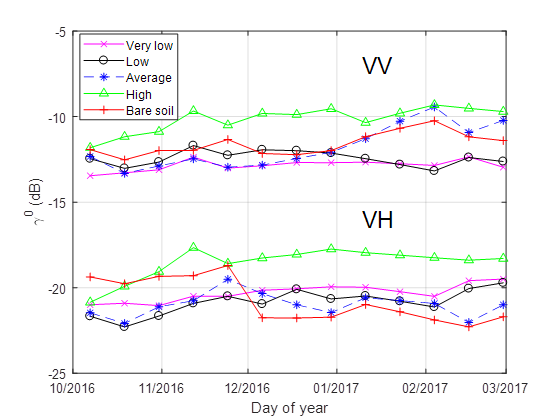}
\caption{The temporal profiles of the five different classes w.r.t. the polarization VV (top) and the polarization VH (bottom). \label{fig:sarSignature}}
\end{figure}

The Sentinel-1 single look complex  dataset includes 13 acquisitions in TOPS mode from 07 October 2016 to 28 February 2017, with a temporal baseline of 12 days. This is dual-polarization (VV+VH) data, resulting in 26 images. Figure \ref{fig:sarSignature} summarizes the temporal profiles of the five winter vegetation quality classes per polarization. Each time series has 13 points (one for each acquisition in TOPS mode) ordered considering the temporal dimension.

\subsection{Pre-procesing}

First, a master image was chosen and all images  were coregistered with taking into account of TOPS mode to the  master image \cite{6130599}.  Five-look (5 range looks) intensity  images were generated and radiometrically calibrated for range spreading loss, antenna gain, normalized reference area and the calibration constant that depends on the parameters Sentinel-1 SAR header.  

\subsection{Temporal filtering}

Reliable estimates of the intensity from a distributed target require that the estimated number of looks  is sufficiently large. Speckle filtering is often used to increase the ENL with loss of spatial resolution \cite{Quegan01}. In properly coregistered multitemporal datasets it is possible to employ the technique of temporal filtering, which in principle increases radiometric resolution without degrading spatial resolution.  The temporally filtered images usually are markedly diminished speckle with little or no reduction in spatial resolution. In this paper, We improve the time series SAR Sentinel-1 dataset by  exploiting a temporal filtering developed by \cite{Quegan01} to reduce noise while retaining as much as possible the fine structures present in the images. 

\subsection{Geocoding}

After pre-processing and filtering all the processed images
were in the imaging geometries of the master image. In order to create a unified dataset all image data had to be orthorectified into map coordinates. This was accomplished by creating a simulated SAR image from a SRTM DEM 30m, and using the simulated SAR image to coregister the two image sets. The pixel size of the orthorectified image data is 10 m. After geocoding, all intensity images are transformed to logarithm dB scale, normalized to values between 0-255 (8 bits) and inputted into classifiers. The SAR Sentinel-1 data  were processed by the IRSTEA TomoSAR platform, which offers SAR, interferometry and tomography processing  \cite{DinhLPS2016}.

\section{Recurrent Neural Network\label{sec:RNN}}
\begin{figure}[ht!]
\centering
\includegraphics[height=0.1\textwidth,width=0.30
\textwidth]{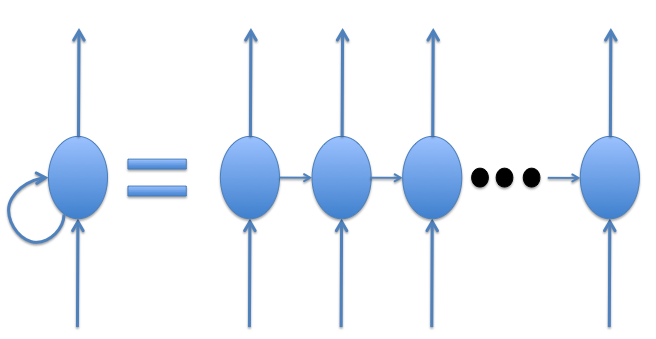}
\caption{RNN Unit (on the left) and unfolded structure (on the right). \label{fig:RNN}}
\end{figure}

Recurrent Neural Networks are well established machine learning techniques that demonstrate their quality in different domains such as speech recognition, signal processing, and natural language processing \cite{SomaMSFN15,LinzenDG16}. Unlike standard feed forward networks (i.e., Convolutional Neural Networks (CNNs)), RNNs explicitly manage temporal data dependencies since the output of the neuron at time t-1 is used, together with the next input, to feed the neuron itself at time t. A sketch of a typical RNN neuron is depicted in Fig. \ref{fig:RNN}. 

Among the different RNN models, Long-Short Term Memory~\cite{HochreiterS96} (LSTM) and Gated Recurrent Unit~\cite{ChoMGBBSB14} (GRU) are the two most well known RNN units. The main difference between them is related to the number of parameters to learn. Considering the same size of the hidden state, the \textit{LSTM} models has more parameters w.r.t. the \textit{GRU} unit to set.

In the following we briefly describe the two RNN units.
For each of them we supply and discuss the equations that describe its inner behavior. The $\odot$ symbol indicates an element-wise multiplication while $\sigma$ and $\tanh$ represent Sigmoid and Hyperbolic Tangent function, respectively.
The input of a RNN unit is a sequence of variables ($x_1$, ..., $x_N$) where a generic element $x_t$ is a feature vector and $t$ refers to the corresponding timestamp.

\subsection{Long-Short Term Memory\label{sec:IRF} }

The \textit{LSTM} model was mainly introduced with the purpose to learn long term dependencies~\cite{HochreiterS96}, since previous RNN models failed in this task due to the problem of vanishing and exploding gradients. The equations (1), (2), (3), (4), (5) and (6) formally describes the LSTM neuron. 
The LSTM unit is composed of two cell states, the memory $C_t$ and the hidden state $h_t$, and three different gates:  input ($i_t$), forget ($f_t$) and output ($o_t$) gate that are employed to control the flow of information. All the three gates combine the current input $x_t$ with the hidden state $h_{t-1}$ coming from the previous timestamp.
The gates have also two important functions: i) they regulate how much information have to be forgotten/remembered during the process; ii) they deal with the problem of vanishing/exploding gradients. We can observe that the gates are implemented by a sigmoid. This function returns values between 0 and 1.
The LSTM unit also uses a temporary cell state $y_t$ that rescales the current input. 
This temporary cell is implemented by an hyperbolic tangent function that returns values between -1 and 1. Both sigmoid and hyperbolic tangent are applied element-wise.

$i_t$ regulates how much of the current information needs to be maintained ($i_t \odot y_t$) while $f_t$ indicates how much of the previous memory needs to be retained at the current step ($f_t \odot c_{t-1}$). Finally, $o_t$ impacts on the new hidden state $h_t$ deciding how much information of the current memory will be outputted to the next step. The different $W_{**}$ matrices and bias coefficients $b_{*}$ are the parameters learned during the training of the model. Both, the memory $C_t$ and the hidden state $h_t$ are forwarded to the next time step. 

\begin{align*}
i_{t} &= \sigma(W_{ix} x_{t} + W_{ih} h_{t-1} + b_i  ) \tag{1}\\
f_{t} &= \sigma(W_{fx} x_{t} + W_{fh} h_{t-1} + b_f  ) \tag{2}\\
y_{t} &= \tanh(W_{yx} x_{t} + W_{yh} h_{t-1} + b_y  )  \tag{3}\\
c_t &= i_t \odot y_t + f_t \odot c_{t-1}  \tag{4}\\
o_{t} &= \sigma(W_{ox} x_{t} + W_{oh} h_{t-1} + b_o  ) \tag{5}\\
h_t &= o_t \odot \tanh(c_t) \tag{6}
\end{align*}

\subsection{Gated Recurrent Unit}

In~\cite{ChoMGBBSB14} the authors design a new RNN unit with the goal to be much simpler to compute and implement of the \textit{LSTM} model.
The equations (7), (8) and (9) formally describes the \textit{GRU} neuron. 
This unit follows the general philosophy of the \textit{LSTM} model implementing gates and cell states but, conversely to the \textit{LSTM} model, the \textit{GRU} unit has only two gates: update ($z_t$) and reset ($r_t$) gates and one cell state: the hidden state ($h_t$). 
Also in this case, the two gates combine the current input ($x_t$) with the information coming from the previous timestamps ($h_{t-1}$).
The update gate effectively controls the trade off between how much information from the previous hidden state will carry over to the current hidden state and how much information of the current timestamps need to be kept. This acts similarly to the memory cell in the \textit{LSTM}
unit supporting the RNN to remember longterm information.

On the other hand, the reset gate monitors how much information of the previous timestamps needs to be integrated with the current information. 
As each hidden unit has separate reset and update gates, each hidden unit will learn to capture dependencies over different time scales. Those
units that learn to capture short-term dependencies will tend to have reset gates that are frequently active, but those that capture longer-term dependencies will have update gates that are mostly active~\cite{ChoMGBBSB14}.

\begin{align*}
z_{t} &= \sigma(W_{zx} x_{t} + W_{zh} h_{t-1} + b_z  ) \tag{7} \\
r_{t} &= \sigma(W_{rx} x_{t} + W_{rh} h_{t-1} + b_r  ) \tag{8} \\
h_{t} &= z_t \odot h_{t-1} +  \tag{9} \\
& (1-z_t) \odot \tanh( W_{hx} x_{t} + W_{hr} (r_t \odot h_{t-1})+b_h  )
\end{align*}

\subsection{RNN-Based Time Series Classification\label{sec:IRF-3} }
To perform the classification task, for each of the RNN unit, we build a deep architecture stacking together five units. The use of multiple units, similarly to what is commonly done for CNN networks combining together several convolutional layers~\cite{BengioCV13}, will allow to extract high-level non-linear temporal dependencies available in the remote sensing time series.

The RNN model learns a new representation of the input sequences but it does not make any prediction by itself. To this end, a SoftMax layer~\cite{Graves13} is stacked on top of the last recurrent unit to perform the final multi-class prediction. 
The SoftMax layer has as many neurons as the number of the classes to predict. We choose the SoftMax instead of the Sigmoid function because the value of the SoftMax layer can be seen as a probability distribution over the classes that sum to 1 while each of the Sigmoid neurons can output a value between 0 and 1. This is due to the fact that, for the SoftMax neuron, the values are normalized per layer while no normalization is performed in the case of Sigmoid layer. This is why, in our context (multi-class prediction), we prefer the SoftMax instead of Sigmoid layer since we know that our samples exclusively belong to a single class.
From an architectural point of view, the connection between the RNN model and the SoftMax layer is realized fully connecting the last hidden state vector of the recurrent architecture with the SoftMax neurons. This schema is instantiated for both \textit{LSTM} and \textit{GRU} units coming up with two different classifiers: an \textit{LSTM}-based and a \textit{GRU}-based classification schema.


\section{Experimental Results\label{sec:Results}}

\subsection{Experimental Settings }
We compare the LSTM and the GRU-Based Time Series models with standard machine learning approaches,  Random Forest (RF) and Support Vector Machine (SVM), usually employed in the remote sensing field \cite{Flamary15}. 

For the \textit{RF} model, we set the number of generated trees equals to 400 and we allow a maximum tree depth of 25. For the \textit{SVM} model we use RBF kernel with default gamma and complexity parameter equals to 10$^6$. For Random Forest we used the python implementation supplied by the Scikit-learn library  \cite{scikit-learn} while for \textit{SVM} we use the LibSVM implementation \cite{CC01a}. 

Considering the RNN-Based classifiers, we set the number of hidden dimensions equals to 512. An initial learning rate of $5 \times 10^{-4}$ and a decay of $5 \times 10^{-5}$ is employed. We have implemented the model via the \textit{Keras} python library  \cite{chollet2015} with \textit{Theano} as back end. 
To train the model we have used the \textit{Rmsprop} strategy that is a variant of the Stochastic Gradient Descent \cite{DauphinVCB15}. The loss function being optimized is the categorical cross-entropy that is the standard loss function employed for multi-class classification tasks ~\cite{Zhang16}. The model is trained for 350 epochs with a batch size equals to 64. To validate the different methods, we perform a 5-fold cross validation on the dataset as shown in Tab. \ref{Table:occSolDistrib}.  In order to assess classification performances, we use not only the Global Accuracy and Kappa measures but also average and per-class F-Measure. Experiments are carried out on a workstation Intel(R) Xeon(R) CPU E5-2667 v4@3.20Ghz with 256 GB of RAM and GPU TITAN X.

\subsection{Results and Discussions}
Table ~\ref{Table:VHVV} summarizes the results of the different classification approaches on the SAR Sentinel 1 Time Series data considering the winter vegetation quality coverage task. 
At first glance, we can observe a significant performance gain between RNN-based classification approaches and classical Machine Learning methods (\textit{RF} and \textit{SVM}). The performance gain involves all the different evaluation metrics. In average, both \textit{LSTM} and \textit{GRU} improve the classification performances of more than 7 point of accuracy/F-Measure. Between the two RNNs models, the \textit{GRU}-based method obtains slightly better results than the \textit{LSTM} one.

\begin{figure}[!ht]
\centering
\includegraphics[scale=0.7]{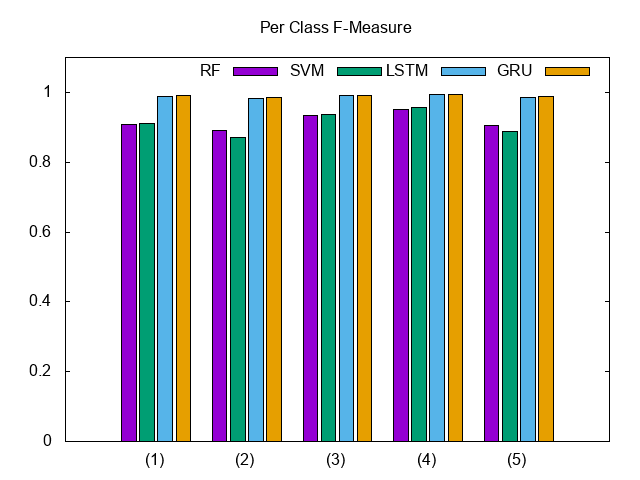}
\caption{Per Class F-Measure of the different approaches. 
\label{fig:perClassF}}
\end{figure}

Figure \ref{fig:perClassF} supplies an F-Measure per class comparison. Here we can have a more precise comprehension of the behavior of the different methods. We can see that the gain supplied by the RNN-based methods involve all the five classes resulting in equally good results on all the them. Conversely, we can note that both \textit{RF} and \textit{SVM} have different behaviors considering different classes. Both classifiers obtain the best performances on the \textit{High} class (4) and the lowest performances on the \textit{Low} class (2). 
This behavior can be explained considering the temporal profiles of both VV and VH presented in Fig.  \ref{fig:sarSignature}. The \textit{Strong} class (depicted with green lines in Fig.  \ref{fig:sarSignature}) has a clear and distinct profile, this facilitate its detection without necessity to consider its temporal correlation. On the other hand, the \textit{Low} class (depicted with black lines in Fig. \ref{fig:sarSignature}) intersects the temporal profiles of all the other classes multiple times. This is probably why, both \textit{RF} and \textit{SVM} approaches are not capable to correctly detect this class since they ignore the temporal correlation of the data. 
On the other hand we can observe that the RNN-Based approaches well discriminate among all the classes since they extract and summarize the important signal portions that support the discriminative task among the different winter quality coverage classes.

\begin{figure}[!ht]
\centering
\subfigure[\label{rf}]{\includegraphics[scale=0.5]{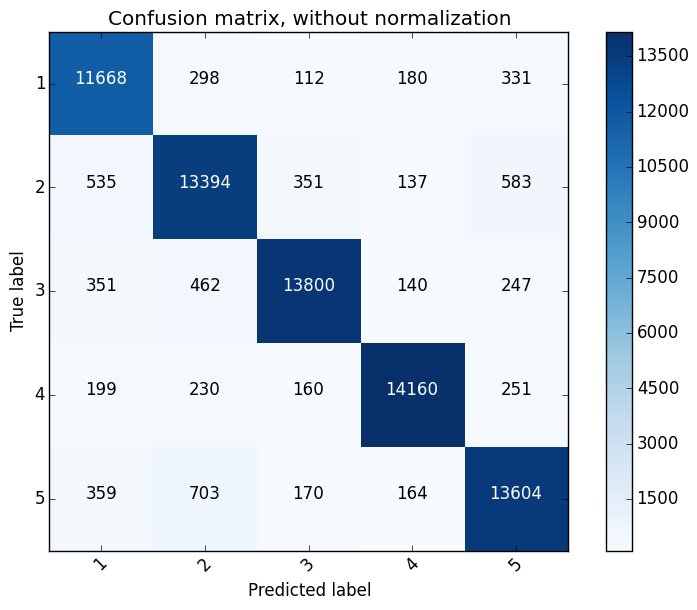}}
\subfigure[\label{svm}]{\includegraphics[scale=0.5]{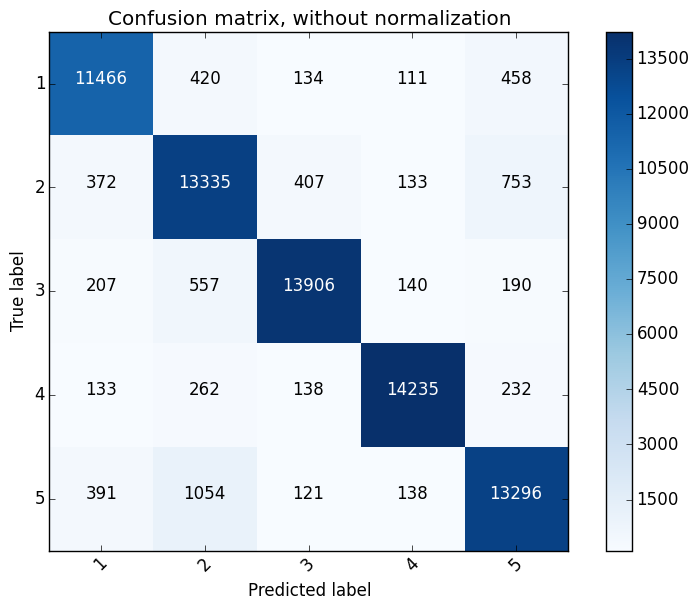}}
\subfigure[\label{lstm}]{\includegraphics[scale=0.5]{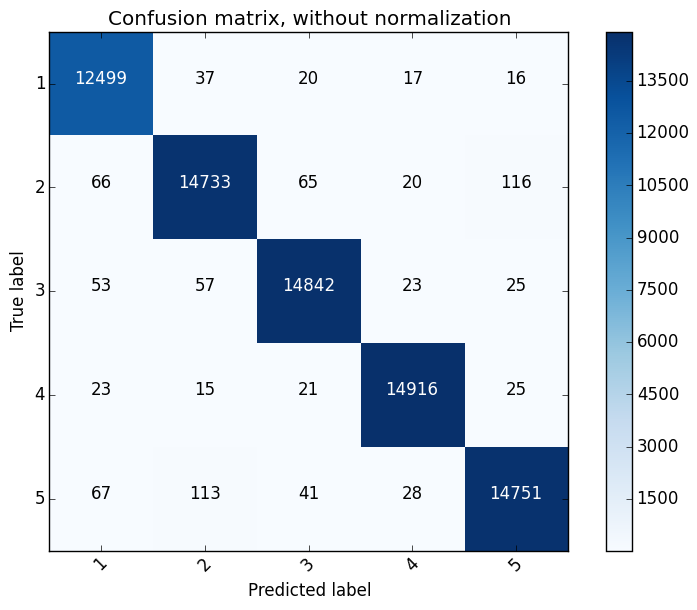}}
\subfigure[\label{gru}]{\includegraphics[scale=0.5]{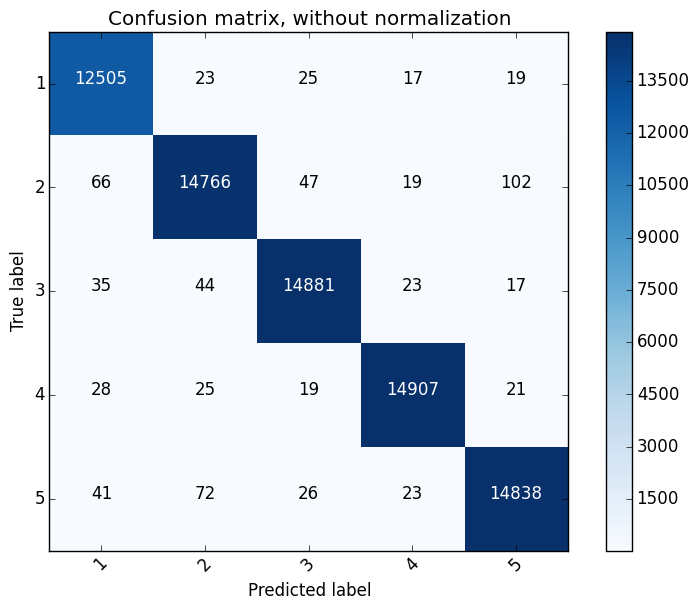}}
\caption{ Confusion Matrices on the SAR SENTINEL 1 Time Series Data of the different approaches: a) \textit{RF}, b) \textit{SVM}, c) \textit{LSTM} and d) \textit{GRU}. \label{fig:ContiTables} }
\end{figure}

Figure \ref{fig:ContiTables} shows the Confusion Matrices for each method. We can observe that the sum of the counts of a matrix is equal to the number of pixels we have in our dataset. To produce the Confusion Matrices, for each classifier, we have aggregated the predictions over the 5-folds. 
We can observe that an heavy misclassification rate happens between the \textit{Low} (2) and \textit{Bare soil} (5) classes. This is true for all the different classifiers. However, for the RNN-based approaches this misclassification error is not so high. Conversely, in the case of both \textit{RF} and \textit{SVM} this misclassification behavior is critical. Another problem we can observe is related to the discrimination between the \textit{Low} and the \textit{Very Low} classes. Also in this case, the standard machine learning approaches suffer and they incur in an high misclassification rate.

The joint optimization of non linear input transformations along with the classifier, proper to all deep learning approaches, provides a valuable strategy to discriminate among the different winter vegetation quality classes. Furthermore, as expected, the ability of RNNs to deal with the temporal correlations, characterizing the SAR Sentinel-1 data, results in a gain of performance on all classes with particular emphasis on such classes that share similar temporal behaviors. All these results indicate that the RNN models (both \textit{LSTM}-based and \textit{GRU}-based) are well suited to detect and exploit temporal dependencies as opposed to common classification approaches that do not explicitly leverage temporal correlations.

\begin{table}[tp]
\centering
\begin{tabular}{|l|c|c|} \hline
\textbf{ID} & \textbf{Vegetation quality class} & {Number of pixels} \\ \hline
(1) & Very low & 12\,589 \\ \hline
(2) & Low & 15\,000 \\ \hline
(3) & Average & 15\,000 \\ \hline
(4) & High & 15\,000 \\ \hline
(5) & Bare soil & 15\,000 \\ \hline
\end{tabular}
\caption{Distribution of the number of pixel per class. \label{Table:occSolDistrib}}

\end{table}

\begin{table}[tp]
\centering
\begin{tabular}{|c|c|c|c|}
\hline 
Classifier & F-measure & Accuracy & Kappa\tabularnewline
\hline 
\hline 
\textit{RF} & 91.77\% & 91.79\% & 0.897 \\
\hline 
\textit{SVM} & 91.22\% & 91.25\% & 0.890 \\
\hline 
\textit{LSTM} & 98.83\% & 98.83\% & 0.985\\
\hline 
\textit{GRU} & \textbf{99.05}\% & \textbf{99.05}\% & \textbf{0.988} \\
\hline 
\end{tabular}
\caption{5-Fold cross valid on the time series SAR Sentinel-1 data. \label{Table:VHVV}
}
\end{table}

\section{Conclusions\label{sec:conclusion}}

In this letter, we demonstrate the capabilities of SAR Sentinel-1 and deep recurrent neural network concerning the mapping winter vegetation quality coverage. We assess the benefit of using the \textit{LSTM} and the \textit{GRU} RNNs to perform winter vegetation quality coverage prediction on a time series of SAR Sentinel-1 images. 
The performances of RNNs clearly surpass the results of classical machine learning approaches commonly used in the remote sensing. The experiments highlight the appropriateness to use deep learning models (RNNs) that explicitly consider the temporal correlation of the SAR data in order to discriminate between classes that exhibit complex temporal behaviors.

\section{Acknowledgments}

The authors wish to thank ESA and CNES/TOSCA for supporting this work.
They also thank the Syndicat des Eaux de la Charente Maritime, the water authority of the City of La Rochelle who carried out field surveys, and the Re-Sources program.
The authors acknowledge the National Research Agency in the framework of the program "Investissements d'Avenir" for the GEOSUD project (ANR-10-EQPX-20).

\bibliographystyle{IEEEtran}
\bibliography{paper.bib}

\end{document}